\documentclass[sigconf]{acmart}

\AtBeginDocument{%
  \providecommand\BibTeX{{%
    \normalfont B\kern-0.5em{\scshape i\kern-0.25em b}\kern-0.8em\TeX}}}

\usepackage{booktabs} 

\usepackage{bm}
\usepackage{subfigure}
\usepackage{float}
\usepackage{multirow}
\usepackage{multicol}
\usepackage{amsfonts}
\usepackage{amsmath}
\usepackage{amssymb}
\usepackage{balance}

\copyrightyear{2020} 
\acmYear{2020} 
\setcopyright{acmcopyright}\acmConference[MM '20]{Proceedings of the 28th ACM International Conference on Multimedia}{October 12--16, 2020}{Seattle, WA, USA}
\acmBooktitle{Proceedings of the 28th ACM International Conference on Multimedia (MM '20), October 12--16, 2020, Seattle, WA, USA}
\acmPrice{15.00}
\acmDOI{10.1145/3394171.3413722}
\acmISBN{978-1-4503-7988-5/20/10}

\begin{document}
\fancyhead{}
\title{PCPL: Predicate-Correlation Perception Learning \\ for Unbiased Scene Graph Generation}

\author{Shaotian Yan$^{1,*}$, \ \ Chen Shen$^{2,{\dagger}}$, \ \ Zhongming Jin$^{2}$, }
\author{Jianqiang Huang$^{2}$, \ \ Rongxin Jiang$^{1,3}$, \ \ Yaowu Chen$^{1,4,{\dagger}}$, \ \ Xian-Sheng Hua$^{2}$ }

\affiliation{%
\institution{$^{1}$ Zhejiang University \ \ $^{2}$ DAMO Academy, Alibaba Group}
}
\affiliation{\institution{$^{3}$ Zhejiang University Embedded System Engineering Research Center, Ministry of Education of China}}
\affiliation{\institution{$^{4}$ Zhejiang Provincial Key Laboratory for Network Multimedia Technologies}}
\affiliation{\{yanshaotian, huaxiansheng\}@gmail.com, \ \ \{jason.sc, zhongming.jinzm, jianqiang.hjq\}@alibaba-inc.com}

\renewcommand{\shortauthors}{Yan et al.}
\renewcommand{\thefootnote}{\fnsymbol{footnote}}
\begin{abstract}
  Today, scene graph generation (SGG) task is largely limited in realistic scenarios, mainly due to the extremely long-tailed bias of predicate annotation distribution.
  Thus, tackling the class imbalance trouble of SGG is critical and challenging. 
  In this paper, we first discover that when predicate labels have strong correlation with each other, prevalent re-balancing strategies (e.g., re-sampling and re-weighting) will give rise to either over-fitting the tail data (e.g., \emph{bench sitting on sidewalk} rather than \emph{on}), or still suffering the adverse effect from the original uneven distribution (e.g., aggregating varied \emph{parked on/standing on/sitting on} into \emph{on}). 
  We argue the principal reason is that re-balancing strategies are sensitive to the frequencies of predicates yet blind to their relatedness, which may play a more important role to promote the learning of predicate features.
  Therefore, we propose a novel Predicate-Correlation Perception Learning (PCPL for short) scheme to adaptively seek out appropriate loss weights by directly perceiving and utilizing the correlation among predicate classes. 
  Moreover, our PCPL framework is further equipped with a graph encoder module to better extract context features. Extensive experiments on the benchmark VG150 dataset show that the proposed PCPL performs markedly better on tail classes while well-preserving the performance on head ones,  which significantly outperforms previous state-of-the-art methods.
\end{abstract}

\ccsdesc[500]{Computing methodologies~Scene understanding}

\keywords{Scene Graph Generation, Long-tailed Bias, Correlation Perception}

\maketitle

{\small\textbf{ACM Reference Format:}\\
Shaotian Yan, Chen Shen, Zhongming Jin,  Jianqiang Huang,  Rongxin  Jiang,  Yaowu Chen,  Xian-Sheng Hua. 2020. PCPL: Predicate-Correlation  Perception Learning for Unbiased Scene Graph Generation. In  \emph{Proceedings of the 28th ACM International Conference on  Multimedia (MM 2020), October 12--16, 2020, Seattle, WA, USA}. ACM,  New York, NY, USA, 9 pages. \\ https://doi.org/10.1145/3394171.3413722}

\footnotetext[1]{This work was done during research intern at Alibaba.}
\footnotetext[2]{Corresponding authors.}

\begin{figure}
    \centering
    \includegraphics[scale=0.5]{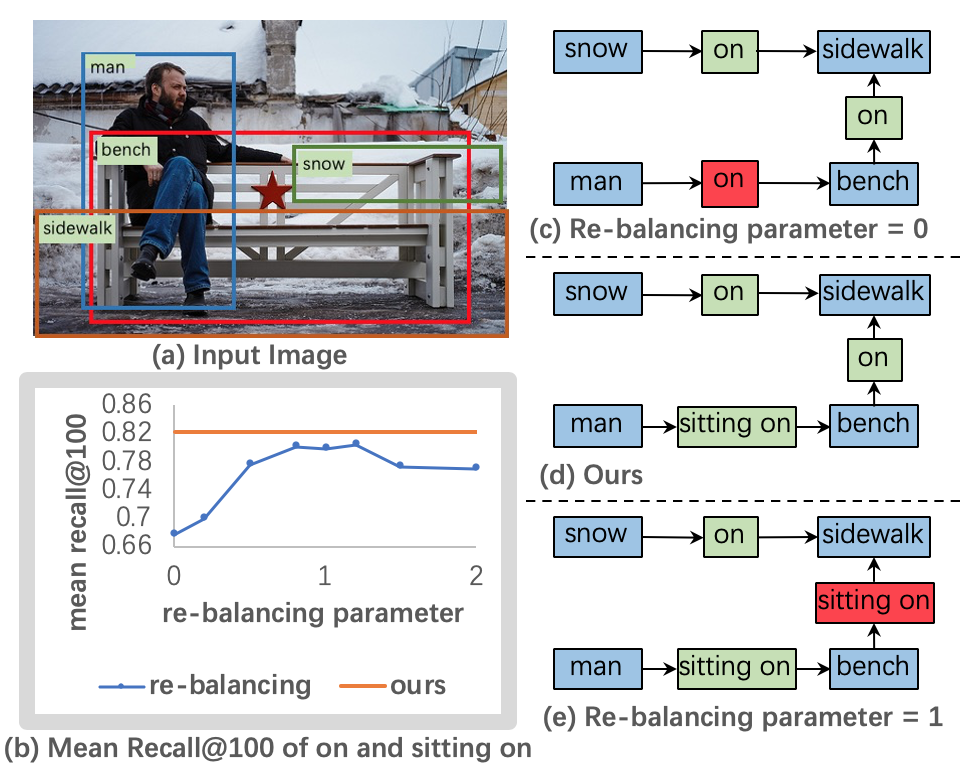}
    \caption{A comparison of re-balancing(e.g., re-weighting) methods and our PCPL. The training weight of re-balancing is set to nth power of the inverse of class sample frequencies where n is the value of re-balancing parameter. (a) An input image with bounding boxes and object labels. (b) The blue curve depicts the mean recall@100 of re-balancing for "on" and "sitting on" under different settings of parameter while the orange curve indicates the performance of our PCPL. It should be noted that 
    there is no intersection between the PCPL and re-balancing training process. (c) SGG with re-balancing parameter = 0. (d) SGG from the proposed PCPL. (e) SGG with re-balancing parameter = 1. Red boxes in (c) and (e) denote wrong predication.}
    \label{fig:1}
\end{figure}

\section{Introduction}

Scene graph generation (SGG)\cite{scene_graph}, which is a visual task to detect objects and recognize semantic relationships between different objects in an image, can serve as a powerful structural representation of images and benefit other high-level Vision-and-Language tasks such as image generation\cite{johnson18,tripathi19,yikang19}, image retrieval\cite{scene_graph,wang20,ramnath19,schroeder20}, visual question answering\cite{yang18,ghosh19,lee19} and image captioning\cite{li19,gu19,yang19}. Taking advantage of the remarkable feature representations of convolutional neural networks (CNNs)\cite{CNN} and diverse contextual feature fusion strategies (e.g., message passing\cite{xu17,li18}, lstm\cite{zellers18}), a variety of methods have made significant progress to improve the recall evaluation metric performance of SGG tasks. Some other works\cite{chen19,tang19} further utilize the co-occurring 
language regularity of typical subject-predicate-object relationship triplets as prior information to enhance overall performance. 
However, in practice the SGG benchmark datasets such as Visual Genome 150 (VG150)\cite{krishna17,xu17} always have extremely long-tailed predicate label distributions (i.e., imbalanced annotation bias in training data, dominating by a few classes which occupy most of the data). Although achieving encouraging performance on head classes (e.g., on/has), these previous efforts are not feasible to obtain outstanding accuracy on predicting fewer but more meaningful predicate samples(e.g., sitting on/riding/looking at/eating/parked on), making them largely limited for supporting high-level tasks in real-world scenarios.

Prevalent class re-balancing strategies are introduced into SGG recently, examined by \citet{tang20} at training stage in order to tackle the challenging long-tailed training data bias problems. In general, the prominent class re-balancing methods are roughly summarized as two types, which are adjusting the sample proportion within a mini-batch (i.e., re-sampling) or assigning relatively higher costs to tail samples (i.e., re-weighting). These two categories share the same connotation of manually tuning sampling frequencies or classifier weights based on the numbers of different class samples during training process to simulate the test distributions. 
These effective strategies indeed promote the overall mean recall evaluation metric for SGG benchmark datasets, however, when  
going deeper to examine specific predicate cases, we unexpectedly find that the performance is not satisfactory under the circumstances that predicate labels have strong correlation with each other.
Fig.~\ref{fig:1} comprehensively illustrates our observation. Taking a semantically closely related predicate pair --- \emph{on} (i.e., head class, occupying a large proportion of annotations) and \emph{sitting on} (i.e., tail class, having rarely few samples) --- as example, it can be seen from Fig.~\ref{fig:1} that, class re-balancing strategies, which merely rely on the manual tuned classifier weights based on the numbers of samples, give rise to either over-fitting the tail data when re-balancing parameter is relative high (\emph{bench sitting on sidewalk} rather than \emph{on}, shown in Fig.~\ref{fig:1}(e)) or still suffering the side effect from the original uneven distribution when re-balancing parameter is relative low (aggregating \emph{man sitting on bench} into \emph{man on bench}).
As shown in Fig.~\ref{fig:1}(b), the optimal point that maximizes the recall of both classes is hardly reachable by manually tuning the re-balancing parameter. We argue the principal reason is that re-balancing strategies merely utilize the frequencies of classes yet neglect their semantic relatedness, which may play a more important role to catalyze the learning of predicate features. To the contrary of other classification tasks, SGG essentially involves complex semantic correlations among different ground truth predicate annotations, which are insensitive to the class frequencies. 

Consequently, we naturally put forward an assumption that the performance of predicates having strong correlations with multi classes will benefit from the learning of correlated ones, as a consequence of which, smaller loss weights are acceptable, otherwise those predicates tend to dominate other classes by severely degrading their recall, and vice versa. In view of that, we propose a novel Predicate-Correlation Perception Learning (PCPL for short) scheme aiming to tackle the class imbalance trouble of SGG, having the benefit of adaptively seeking out optimal loss weights by directly perceiving and explicitly utilizing the implicit correlations among predicates. Equipped with PCPL, the model is able to markedly improve the predicting results on tail classes and well preserve the performance on head predicates simultaneously, thus the optimum mean recall can be obtained as illustrated in Fig.~\ref{fig:1}(b). 
Specifically, we construct an iteratively updated class graph to perceive the correlations between predicates and the loss weights of classes are appropriately inversed with their relatedness derived from the graph.

Morever, we propose a graph encoding module (GE for short) to encode global context through a series of stacked encoders in a graph manner. A variety of methods have been adopted by previous works to fuse global context into relationship representations. Dual graph message passing\cite{xu17} is relatively out-of-date while BiLSTM\cite{hochreiter97} employed by Neural Motifs\cite{zellers18} achieves better results yet suffers a drawback that different input orders will bring different results. Without introducing additional information, the performance of GGNN \cite{chen19} is not satisfying. Compared with previous methods, our graph encoding module can better capture the relationships between object classes and benefit from being permutation invariant, thus can obtain more robust contextual features, setting a higher baseline. 

In summary, the contributions of this paper are threefold:
\begin{itemize}
    \item We propose a novel Predicate-Correlation Perception Learning (PCPL for short) scheme that is able to alleviate the long-tailed bias of SGG by directly perceiving and explicitly utilizing the implicit correlations between predicate classes, opening up new ideas to tackle the imbalance issue of SGG or other tasks involving correlated classes.
    \item PCPL can significantly promote the predicting results of tail classes while well preserving the performance of head predicates, by adaptively assigning optimal loss weights, which are appropriately inversed with the degrees of relatedness, to different predicates.
    \item Extensive experiments show the effectiveness of PCPL and demonstrate that PCPL achieves a new state-of-the-art.
\end{itemize}

\section{Related work}
\subsection{Scene Graph Methods}
Reasoning about relationships is the major challenge for generating scene graph. There are mainly two approaches in previous works making efforts to improve the performance of relationship prediction. The first approach is to make better use of visual features. \citet{xu17} finds that relationship prediction can be greatly improved by jointly reasoning with contextual information. Message Passing model proposed by Xu iteratively refines its prediction by passing contextual messages along the topological structure of a scene graph. \citet{zellers18} emphasizes the importance of context by introducing BiLSTM to encode global context that can directly inform the local predictors. Second approach is to involve additional information such as semantic labels and statistical correlations to help prediction. \citet{zellers18} embeds GloVe word vectors, statistical correlations of object pairs and relationships to visual features to obtain better results. \citet{chen19} makes further use of statistical correlations. They facilitate scene graph generation by explicitly unifying the statistical knowledge with the architecture of graph neural network.

\citet{chen19,tang19} both take a notice on the class imbalanced issue of SGG by proposing the mean recall@K metric but their works are still confined to better feature extracting. In recent works, the criticalness of long-tailed bias of SGG is addressed \cite{tang20,wen20}. \citet{tang20} employ causal inference in the prediction stage in an effort to remove the training bias while \citet{wen20} make use of a pseudo-siamese network to pursue extracting balanced visual features. 
\begin{figure}[htbp]
\centering
 
\subfigure[]{
    \begin{minipage}[t]{0.5\linewidth}
        \centering
        \includegraphics[scale=0.26]{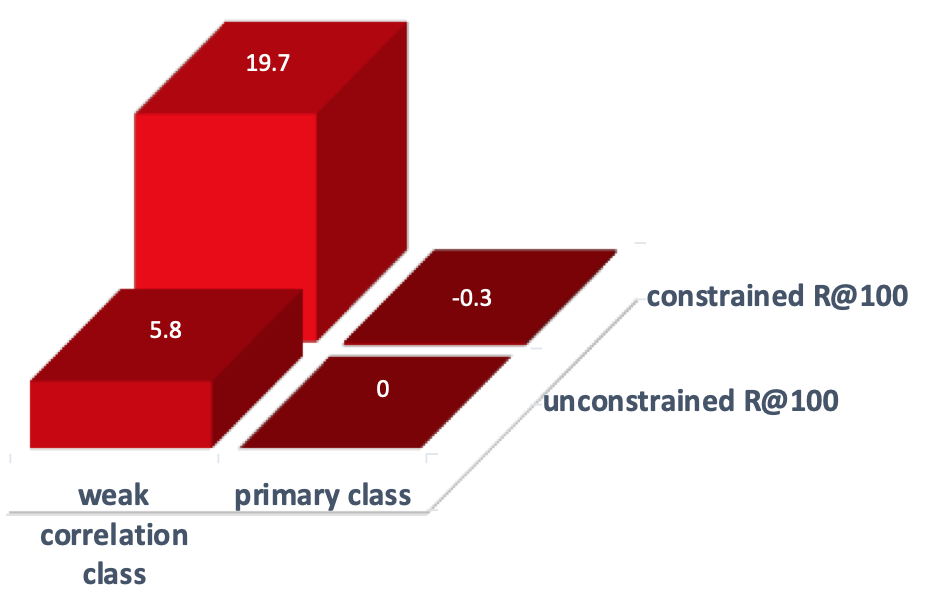}\\
        \vspace{0.02cm}
    \end{minipage}%
}%
\subfigure[]{
    \begin{minipage}[t]{0.5\linewidth}
        \centering
        \includegraphics[scale=0.26]{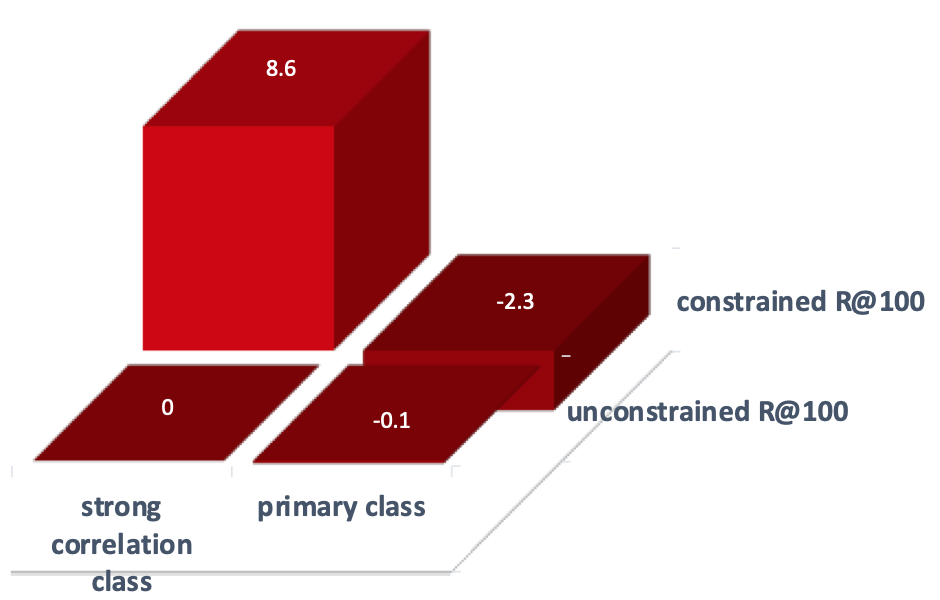}\\
        \vspace{0.02cm}
    \end{minipage}%
}%
 
\centering
\caption{The improvements of constrained and unconstrained recall@100 of re-balancing over cross-entropy in \%. (a) The results of weakly correlated group. (b) The results of strongly correlated group.}
\vspace{-0.2cm}
\label{fig:re-weighting-entropy_compare}
\end{figure}
In contrast, our method perceives and utilizes the implicit correlations among predicate classes based on an innovative observation, aiming to generate unbiased scene graph representations.
\subsection{Class Imbalance}
Real-world large-scale datasets often have long-tailed data distributions. Neural networks trained on these datasets tend to perform poorly on less presented classes. It has become a critical issue for model training. A lot of works has been done to resolve the class imbalance problem. Existing methods can be categorized as re-sampling \cite{shen16,buda18,geifman17,japkowicz02,zou18} and re-weighting\cite{cui19,ren18,khan17}. Re-sampling methods are simple yet effective. They often over-sample (e.g.,\cite{byrd19,ramentol12}) less presented classes or under-sample (e.g.,\cite{lee16,haixiang17}) frequent classes to make the data distribution more balanced. However, they have their downsides. Under-sampling frequent classes will discard a large amount of data, causing waste of data. And it is not practicable when the dataset is extremely imbalanced. Over-sampling less presented classes can lead to over-fitting of the repeatedly sampled classes.
\\\indent Re-weighting methods assign different weights for different classes to balance the loss. The simplest way of re-weighting is to set weights of classes as the inverse of their frequency\cite{huang16,wang17}, but this causes poor performance on frequent classes. \citet{cui19} proposes the definition of effective number of samples and re-weights the loss by the inverse of effective number to address this issue. Another widely used re-weighting method is focal loss proposed by \citet{lin17}. Focal loss down-weights the loss assigned to well-classified examples and focuses training on a sparse set of hard examples.

While most of traditional re-balancing methods merely rely on sample frequencies to manually tune the loss weights or sample ratios of different classes, our proposed method is able to adaptively assign optimal training costs to classes based on their relatedness.
\begin{figure*}[htbp]
\centering
\subfigure[]{
    \begin{minipage}[t]{0.32\linewidth}
        \centering
        \includegraphics[width=1.951in]{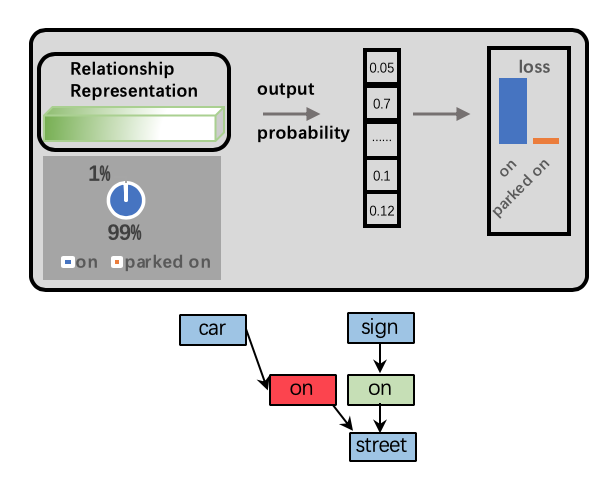}\\
        \vspace{0.02cm}
    \end{minipage}%
}%
\subfigure[]{
    \begin{minipage}[t]{0.3\linewidth}
        \centering
        \includegraphics[width=1.951in]{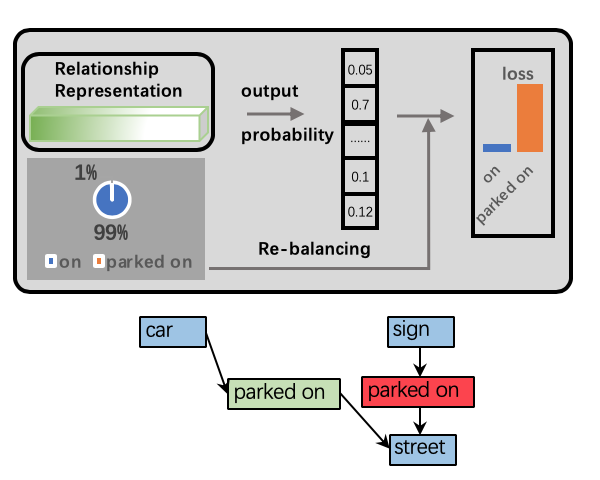}\\
        \vspace{0.02cm}
    \end{minipage}%
}%
\subfigure[]{
    \begin{minipage}[t]{0.3\linewidth}
        \centering
        \includegraphics[width=1.951in]{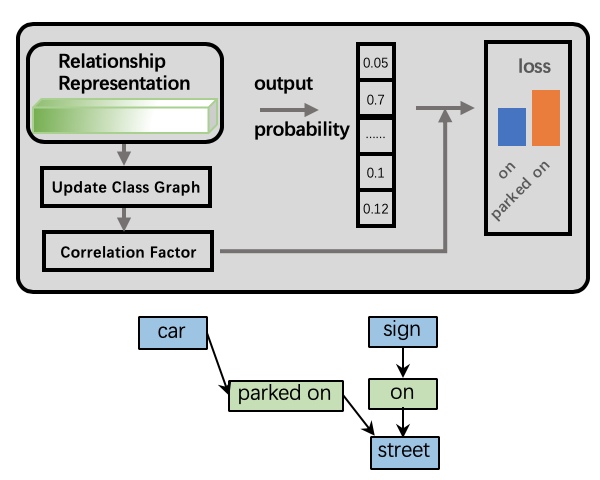}\\
        \vspace{0.02cm}
    \end{minipage}%
}%
                 
\quad
\subfigure[]{
    \begin{minipage}[t]{0.8\linewidth}
        \centering
        \includegraphics[width=3.61in]{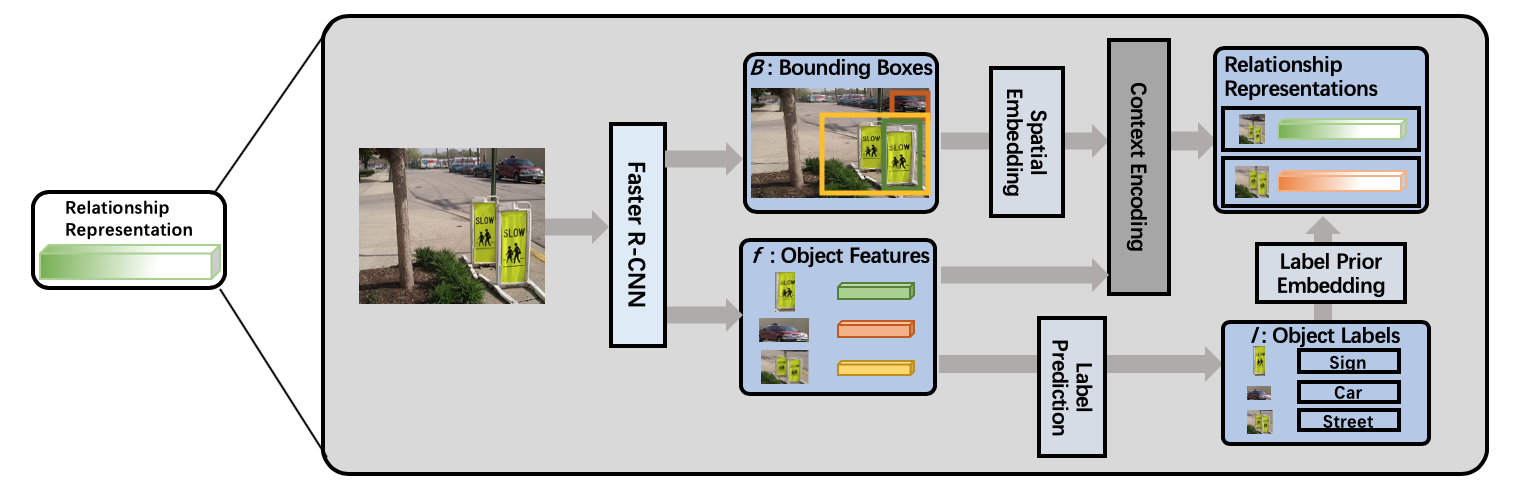}\\
        \vspace{0.02cm}
    \end{minipage}%
}%
\centering
\caption{Illustration of how the baseline method (a), the re-balancing methods (b) and PCPL (c) generate scene graph from relationship representation and the corresponding output scene graph. Red boxes in (a) and (b) denote wrong predictions. (d) The pipeline used to acquire relationship representations. }
\vspace{-0.2cm}
\label{fig:overall_framework}
\end{figure*}
\section{Methods}

\subsection{Problem definition}
Scene graph\cite{scene_graph}, representing a visual scene's detailed semantics, is generated with:
\begin{itemize}
    \item a set of bounding boxes $B = \{{b}_{1}, {b}_{2}, …, {b}_{n}\}$, referring to the spatial locations of detected regions,
    \item a set of labels $O = \{{o}_{1}, {o}_{2}, …, {o}_{n}\}$, containing object label ${o}_{i}$ of the corresponding bounding box ${b}_{i}$,
    \item and $R = \{{r}_{1->2}, {r}_{1->3}, …, {r}_{n->n-1}\}$, denoting the relationships of object pairs.
\end{itemize}
A triplet of a start object $({o}_{i}, {b}_{i})$, an end object $({o}_{j}, {b}_{j})$ and a predicate label ${p}_{i->j}$ connecting the former to the latter make up ${r}_{i->j} \in R$.

 As shown in Fig.~\ref{fig:overall_framework}(d), we conduct a conventional two-stage pipeline which detects the locations and labels of objects first and then outputs the relationship representations. Given an input image containing two strong correlated predicates with great disparity in sample frequencies (i.e.,\emph{parked on} and \emph{on}), the ubiquitous cross entropy loss, employed by most of SGG methods to optimize the framework, causes aggregating \emph{parked on} into \emph{on}, as can be seen in Fig.~\ref{fig:overall_framework}(a). At the other extreme, Fig.~ \ref{fig:overall_framework}(b) illustrates that a typical re-balancing strategy, which assign the 
fixed inverse of frequencies to the sample weights of predicate classes, surprisingly give rise to over-fitting \emph{parked on}. Both strategies fail to achieve satisfactory performance when there exists strong correlation between predicate classes with long-tailed distribution. In view of that, we propose a novel PCPL scheme aiming to tackle the class imbalance trouble of SGG by directly perceiving and explicitly making use of the implicit correlations among predicates. Fig.~\ref{fig:overall_framework}(c) presents an overview of PCPL. We construct an iteratively updated class graph to represent the correlations between predicates. By utilizing the relatedness derived from the graph, PCPL has an 
advantage of adaptively seeking out optimal loss weights instead of manually tuning. Equipped with PCPL, the model is able to markedly improve the predicting results on tail classes and well preserve the performance on head predicates simultaneously. In subsequent sections, we will start with an innovative observation of re-balancing on SGG and then describe our method in detail.

\subsection{An observation of re-balancing}
As mentioned above, we find that re-balancing methods are of no avail when predicates are closely correlated to each other. In an effort to further explore the origins of this phenomenon, this section will discuss the influence of class correlations on the effectiveness of re-balancing for SGG before diving into our method.

Considering that it is hard to judge the degree of correlations by human intuition, we utilize the distance between class centers in feature space as a quantitatively measurement for the relatednesses, as the feature clusters of two correlated classes tend to be closer than two independent classes. Concretely, we employ a learnable variable ${\textbf{v}}_{k}$, jointly trained with the SGG model, to represent the center of predicate class $k$ in the feature space. The dimension of ${\textbf{v}}_{k}$ is the same as the output relationship feature before the last fully connected layer of SGG model. For output features $\{{f}_{1},{f}_{2}...{f}_{N}\}$ and predicate labels $\{{l}_{1},{l}_{2}...{l}_{N}\}$, the loss to update $\textbf{v}$ is defined as:
\begin{equation}\label{equ:Lcenter}
    L_{center}=\frac{1}{N}\sum_{i=1}^N({f}_{i} - {\textbf{v}}_{{l}_{i}})^2
\end{equation}
where $N$ is the count of ground truth predicate annotations in the mini-batch and ${\textbf{v}}_{{l}_{i}}$ is the corresponding center variable for output feature ${f}_{i}$.
Notably, the gradient of $\textbf{v}$ will not pass to feature ${f}$ in backward propagation in order not to mess up the training procedure of SGG model.
\\\indent Directly following the training process, we acquire the correlations between predicate classes:
\begin{equation}\label{equ:G}
    {\textbf{e}}_{kj}=||{\textbf{v}}_{j}-{\textbf{v}}_{k}||_2
\end{equation}
Classes with larger $\textbf{e}$ are more independent while smaller $\textbf{e}$ means stronger correlation. $\textbf{e}$ between class $k$ and it self equals 0.
\\\indent To provide a more intuitive illustration, comparative experiments are designed for two observation predicate groups with opposed level of correlations. Specifically, The first group consists of two predicates, a primary class, occupying a large proportion of annotations, and a strong correlation class, with far fewer samples but closely correlated with the primary one. The same primary class and a weak correlation class, having the same sample frequency with the strong one yet relatively independent, make up the other group. To get rid of the influence of other classes, we remove irrelevant annotations from the dataset for each group separately, thus to make the results clearer and more concise. Here we regard "has", "with" and "looking at" as the primary predicate, strong correlation predicate and weak correlation predicate respectively. 
Obtained from Eq.~\ref{equ:G}, $\textbf{e}$ between \emph{has} and \emph{with} is 4.57 while that between \emph{has} and \emph{looking at} is 20.96, which means that the relatedness between \emph{has} and \emph{with} is strong and that between \emph{has} and \emph{looking at} is weak. For a fair comparison, we randomly down sample the frequency of \emph{with} to the same scale of \emph{looking at}.
\\\indent Examined with a same baseline model on each group, the constrained and unconstrained R@100 improvements of re-weighting over cross-entropy are revealed in Fig.~\ref{fig:re-weighting-entropy_compare}. The results demonstrate that both the constrained and unconstrained R@100 of the weak correlation predicate increase notably with almost no impact on the primary predicate. In the other group, though the constrained R@100 of the strong correlation predicate occurs a minor rise, that of the primary predicate happens a relatively significant decrease, while there is no obvious change on the unconstrained R@100 of both predicates. The contrast results of the two groups indicate that re-balancing, to some extent, is able to alleviate the class imbalance trouble when classes are independent. However, when it comes to classes closely correlated with each other, these strategies, sensitive to class frequency but blind to the correlations between classes, result in over-fitting to tail classes. Scene graph generation task, predicating relationship between instances, involves critically complex correlations between predicates, which fully exposes the shortcoming of re-balancing. In stark contrast, our proposed PCPL scheme can achieve a satisfying performance on both head and tail classes by adaptively assigning optimal training costs, which are appropriately inversed with the degrees of relatedness, to predicates, opening up new ideas to tackle the imbalance issue of SGG or other tasks involving correlated classes. Although predicates having strong correlations with multi classes are assigned with relatively smaller loss weights, their performance will benefit from the learning process of correlated ones, while other predicates gain improvements on account of higher training costs. The detailed process of PCPL will be described in the next section. 
\begin{figure*}[htbp]
\centering
 
\subfigure[]{
    \begin{minipage}[t]{0.166\linewidth}
        \centering
        \includegraphics[width=1.11in]{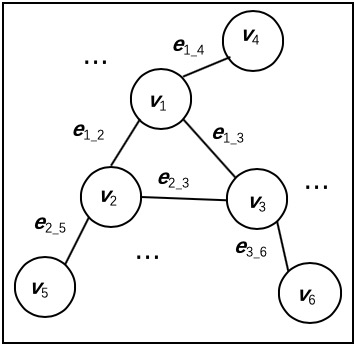}\\
        \vspace{0.01cm}
    \end{minipage}%
}%
\subfigure[]{
    \begin{minipage}[t]{0.166\linewidth}
        \centering
        \includegraphics[width=1.11in]{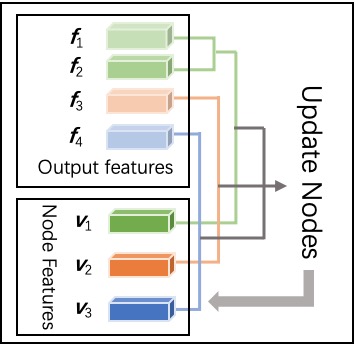}\\
        \vspace{0.01cm}
    \end{minipage}%
}%
\subfigure[]{
    \begin{minipage}[t]{0.166\linewidth}
        \centering
        \includegraphics[width=1.11in]{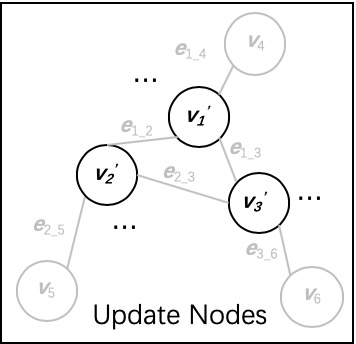}\\
        \vspace{0.01cm}
    \end{minipage}%
}%
%
\subfigure[]{
    \begin{minipage}[t]{0.166\linewidth}
        \centering
        \includegraphics[width=1.11in]{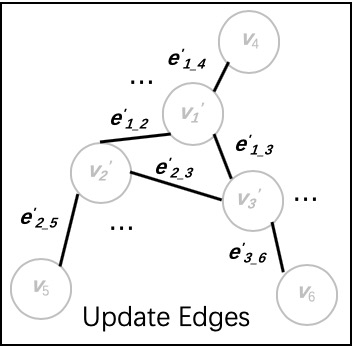}\\
        \vspace{0.01cm}
    \end{minipage}%
}%
\subfigure[]{
    \begin{minipage}[t]{0.166\linewidth}
        \centering
        \includegraphics[width=1.11in]{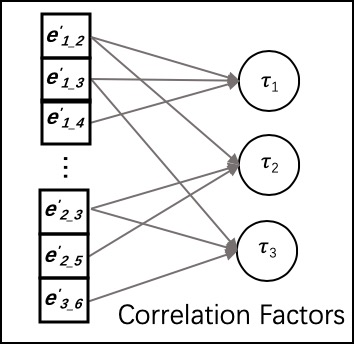}\\
        \vspace{0.01cm}
    \end{minipage}%
}%
\subfigure[]{
    \begin{minipage}[t]{0.166\linewidth}
        \centering
        \includegraphics[width=1.11in]{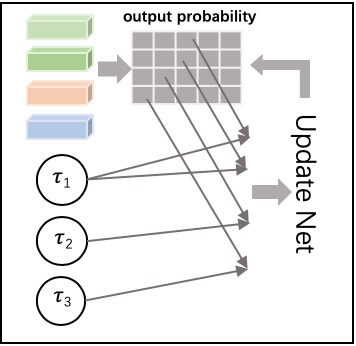}\\
        \vspace{0.01cm}
    \end{minipage}%
}%

\centering
\caption{Illustration of the proposed Predicate-Correlation Perception Learning (PCPL) scheme. (a) A learnable class graph is constructed with each node representing the center of one predicate class and edges representing their correlations.  (b) (c) (d) The graph is jointly trained with SGG model. (e) Correlation factors are derived from the graph. (f) We utilize correlation factors to adaptively assign optimal loss weights to predicate classes.}
\vspace{-0.2cm}
\label{fig:PCPL}
\end{figure*}
\subsection{Learning Process}
The class correlations are dynamically changing along with the optimization of the feature extracting network. For this reason, as is shown in Fig.~\ref{fig:PCPL}, we construct a learnable class graph to dynamically perceive the relatedness between predicates. As the network gradually converges, the graph we built is also achieving a relatively stable state. In this way, we are able to guide the learning of the model throughout the whole training process. The graph consists of a set of nodes and edges connecting every pair of them. Each node represents the center of one predicate class while the edges connecting nodes represent their degrees of correlation. Given output features, we first update the corresponding ${\textbf{v}}_{i}$ using Eq.~\ref{equ:Lcenter} and update the edges with Eq.~\ref{equ:G} afterwards, as presented in Fig.~\ref{fig:PCPL}(b,c,d). Then the global correlations $\textbf{u}_{i}$ of predicate class ${i}$ is defined as:
\begin{equation}\label{eq:factor}
{\textbf{u}}_{i}=\sum_{j=1}^{N}{\textbf{e}}_{ij}^{'}
\end{equation}
where ${N}$ is the number of predicate classes in the dataset and $\textbf{e}_{ij}^{'}$ is the updated value of edge connecting node ${i}$ and node ${j}$.  Following this, we perform a normalization for $\textbf{u}_{i}$ to obtain the correlation factor $\tau$:
\begin{equation}\label{eq:normalize}
    {\tau}_{i}=\frac{{\textbf{u}}_{i}-min({\textbf{u}})+\epsilon}{max({\textbf{u}})-min({\textbf{u}})}
\end{equation}
where $\epsilon$ denotes a minimal value to prevent ${\tau}_{i}$ from being zero. The correlation factor ${\tau}_{i}$ can be seen as a measure for the independence degree of class ${i}$. After that, ${\tau}_{i}$ is assigned to the classification loss weight of the SGG network, thus to correct the learning process and alleviate the training bias:
\begin{equation}\label{eq:softmax}
    p^{'}_{l_i} = \frac{{e}^{p_{{l}_{i}}}}{\sum_{j=1}^{N}{e}^{p_j}}, 
\end{equation}
\begin{equation}\label{eq:weight}
    {L} = -\sum_{i=1}^N\frac{{\tau}_{{l}_{i}}}{\sum_{k=1}^{{N}_{r}}{({\tau}_{{l}_{k}}})}*\log{p^{'}_{{l}_{i}}}
\end{equation}
where $N_r$ is the count of ground truth predicate classes present at the current mini-batch, $p$ is the probability of each predicate output by the model and $l_i$ is the ground truth label of feature $i$.
\\\indent Moreover, the dynamic graph makes it possible to alleviate the influence of noisy labels. Models are easy to be distracted by noisy labels because their losses are usually higher than normal samples. With the graph, we are able to distinguish and abandon noisy labels, thus to make the learning process more stable to some extent. Given an output feature ${f}_{i}$ with ground-truth label ${i}$, we distinguish whether it is noisy or not by ${D}_{drop}$:
\begin{equation}\label{eq:drop}
    {D}_{drop_j} = {||{f}_{i} - {\textbf{v}}_{i}||_2 - ||{f}_{i} - {\textbf{v}}_{j}||_2} - \frac{{{\textbf{e}}_{ij}}}{\lambda}
\end{equation}
where ${\lambda}$ is a hyper-parameter and ${D}_{drop_j}$ means ${D}_{drop}$ with class ${j}$. Here we set ${\lambda}$ as ${2}$. If any of ${D}_{drop_j}$ is great than zero, we consider ${f}_{i}$ as noisy and abandon the corresponding sample.
\begin{figure}[h]
    \centering
    \includegraphics[scale=0.3]{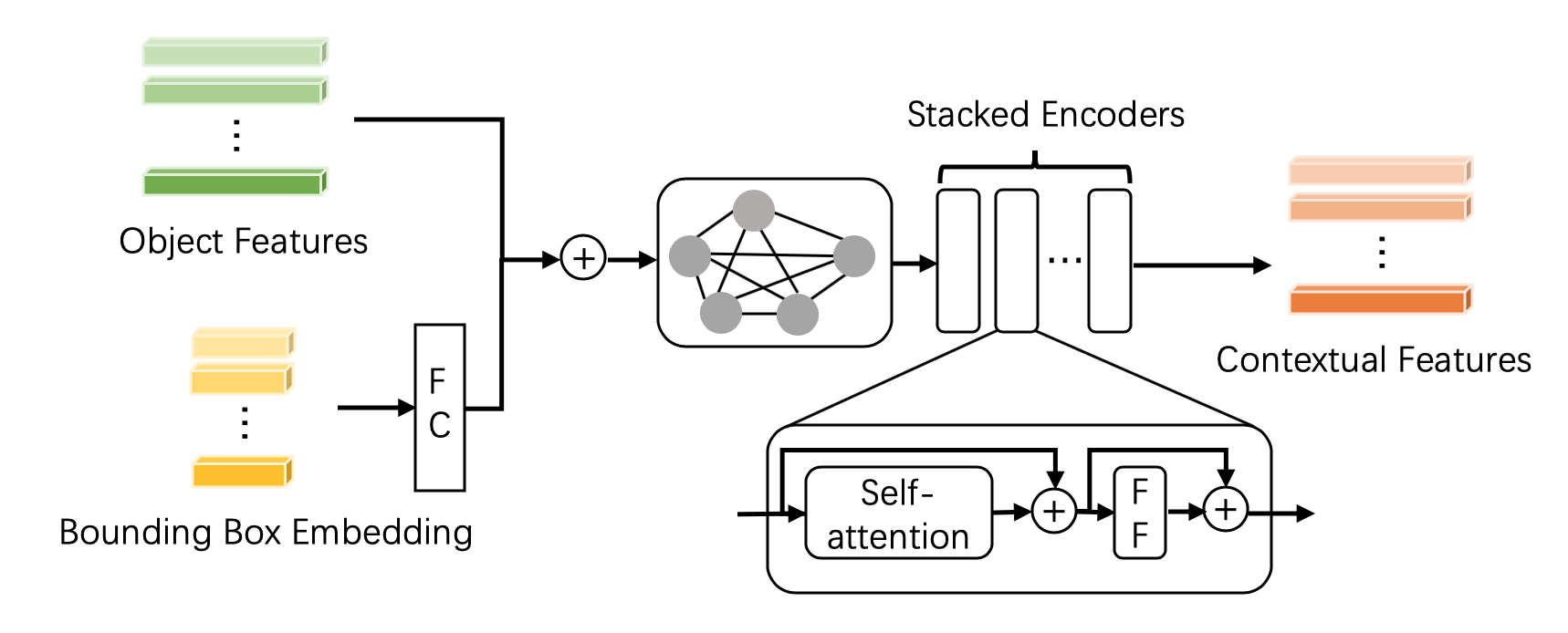}
    \caption{A diagram of the Graph Encoding Module (GE). We fuse object features as well as their corresponding spatial information to construct a graph and obtain contextual features by processing the graph with stacked encoders. Each encoder is permutation invariant by consisting of a self-attention layer and a Feed Forward network (FF).}
    \label{fig:context_encoding}
\end{figure}
\begin{table*}[htbp]
    \centering
    \caption{Performance comparison with state-of-the-art methods on VG150 dataset. The constrained and unconstrained mR@50/100 in \% on PredCls, SGCls and SGGen tasks are presented. As VCTree and TDE do not report the unconstrained mR@K metric, they are not listed in unconstrained results.}
    \begin{tabular}{cccccccccc}
        \toprule
        \multirow{2}{*} & \multirow{2}{*}{Methods} & \multicolumn{2}{c}{PredCls} & \multicolumn{2}{c}{SGCls} & \multicolumn{2}{c}{SGGen} \\
        &  &  mR@50 & mR@100 & mR@50 & mR@100 & mR@50 & mR@100 & Mean\\
        \midrule
        \multirow{5}{*}{unconstrained} & IMP+\cite{xu17} &20.3    &28.9        &12.1     &16.9  &5.4    &8.0      &15.3  \\
        &  FREQ\cite{zellers18}   &24.8   &37.3  &13.5     &19.6  &5.9   &8.9 &18.3  \\
        &  SMN\cite{zellers18}    &27.5   &37.9  &15.4     &20.6  &9.3 &12.9 &20.6  \\
        &  KERN\cite{chen19}   &36.3   &49.0  &19.8     &26.2  & \textbf{11.7}  &\textbf{16.0}  &26.5 \\
        &  \textbf{Ours}   &\textbf{50.6}   &\textbf{62.6}  &\textbf{26.8}    &\textbf{32.8}   &10.4 &14.4    &\textbf{32.9}  \\
        \midrule
        \midrule
        \multirow{7}{*}{constrained} & IMP+\cite{xu17} &9.8    &10.5       &5.8     &6.0  &3.8    &4.8   &6.8  \\
        &  FREQ\cite{zellers18} &13.3   &15.8   &6.8    &7.8    &4.3    &5.6  &8.9  \\
        &  SMN\cite{zellers18}  &13.3   &14.4   &7.1    &7.6    &5.3    &6.1 &9.0  \\
        &  KERN\cite{chen19} &17.7   &19.2   &9.4    &10.0   &6.4    &7.3  &11.7 \\
        &  VCTree\cite{tang20}  &17.9   &19.4   &10.1   &10.8   &6.9  &8.0  &12.2 \\
        &  SMN+TDE\cite{tang20} &25.5  &29.1   &13.1   &14.9   &8.2   &9.8  &16.8 \\
        &  \textbf{Ours}  &\textbf{35.2}   &\textbf{37.8}                &\textbf{18.6}    &\textbf{19.6}  &\textbf{9.5}      &\textbf{11.7}      &\textbf{22.1}  \\
        
        \bottomrule
    \end{tabular}
    \label{tab:state-of-the-art}
\end{table*}
\subsection{Context Encoding}
Given an image $I$, bounding boxes are first detected using Faster R-CNN as described in Fig.~\ref{fig:overall_framework}(d). Besides, for each ${b}_{i}$ in the proposal region set $B$, it also outputs a corresponding feature vector and a possible label ${l}_{i}$, which are of non-context, causing relatively low performance in object and predicate classification. Thus, as shown in Fig.~\ref{fig:context_encoding}, we design a graph encoding module to obtain contextualized representations. Taking the pooled feature vectors as a set of nodes, we use an input network implemented by fully connected layers to expand bounding box coordinates to the same dimension of node features. Following that, we perform an element-wise sum to acquire new representations of nodes, containing spatial information which is also crucial when inferring relationships. Afterwards, we can construct a fully connected undirected graph $G$ by connecting all the nodes together. The edges between nodes represent to what extent nodes can interact with its neighbors. Then we iteratively process the graph with stacked encoders. As Fig.~\ref{fig:context_encoding} illustrates, each encoder consists of a self-attention layer and a Feed Forward network (FF). Every encoder calculates the attention coefficients between nodes and obtains the hidden state of each node by attending over its neighbors:
\begin{equation}
    \hat{\textbf{H}}_{i-1} = \textbf{{H}}_{i-1} + \textbf{Attention}(\textbf{{H}}_{i-1})
\end{equation}
\begin{equation}
    \textbf{H}_{i} = \hat{\textbf{H}}_{i-1} + \textbf{FF}(\hat{\textbf{H}}_{i-1})
\end{equation}
where $\textbf{{H}}_{i}$ is the hidden state of graph $G$ output by $i$th encoder. In this way, messages can be propagated through the whole graph. Eventually, we obtain the final contextual representation of each region after processing several encoders.

\section{Experiments}
\subsection{Experiment Settings}
\subsubsection{Implementation Details}
To keep consistent with previous works \cite{xu17,zellers18,chen19,tang19}, we adopt the Faster-RCNN detector\cite{faster-rcnn}, pretrained on ImageNet\cite{imagenet} and refined on VG150\cite{xu17}, with VGG16\cite{VGG16} being the backbone to generate region proposals. The numbers of stacked encoders and attention heads in graph encoding modules are set to 6 and 12 respectively. All our experiments are conducted using a NVIDIA P100 GPU.

\subsubsection{Dataset}
We evaluate our methods and all the comparison models on Visual Genome\cite{krishna17}, a large-scale dataset commonly used in vision-and-language tasks. Following prior works\cite{xu17,zellers18,chen19,tang19}, we adopt the most popular preprocessed split, VG150\cite{xu17}, which contains the most frequent 150 object categories and 50 predicate classes.
\subsubsection{Evaluation Metrics}
Most of previous works adopt the recall@K (R@K for short) metric that measures the fraction of ground-truth relationship triplets(subject-predicate-object) that appear among the top K most confident predictions in an image\cite{xu17}. However, this metric is easily dominated by a few predicate classes accounting for absolute proportion of data due to the long-tail distribution of annotations. Thus we abandon R@K on most experiments and evaluate all the methods using the mean rcall@K (mR@K for short), proposed by \citet{chen19} and \citet{tang19}, to give a more comprehensive assessment. It is defined as the average R@K of all the predicate classes, which gives a fair performance appraisal for both head and tail classes. Notably, we report R@K in Table 2, which compares different debiasing methods, to avoid over-fitting to tail classes.
Both unconstrained and constrained \cite{zellers18} mR@K are presented on all experiments, which obtained from multi and single output relationships respectively.
\subsubsection{Tasks}
To comprehensively evaluate the performance on different stages of SGG, we adopt the following three tasks:
Predicate classification (PredCls) predicts the predicate classes of a set of given object pairs with ground truth bounding boxes and object labels.
Scene graph classification (SGCls) predicts the object classes for ground truth bounding boxes and predicts the predicate labels of each object pairs.
Scene graph generation (SGGen) only takes the original image as input and sequentially detects the bounding boxes, object labels and then predicts the relationships between object pairs.
\begin{table*}[h]
    \centering
    \caption{Performance comparison with other debiasing methods on VG150 dataset. The R@50/100 and mR@50/100 in \% with and without constraints on PredCls, SGCls and SGGen tasks are presented.}
    \begin{tabular}{ccccccccc}
        \toprule
        \multirow{2}{*} & \multirow{2}{*}{Methods} & \multicolumn{2}{c}{PredCls} & \multicolumn{2}{c}{SGCls} & \multicolumn{2}{c}{SGGen} \\
        &  &  R@50/100 & mR@50/100 & R@50/100 & mR@50/100 & R@50/100 & mR@50/100 & Mean\\
        \midrule
        \multirow{4}{*}{unconstrained}
        & GE + focal loss\cite{lin17}   &\textbf{77.3/85.4}  &26.4/36.2  &\textbf{42.3/46.1}  &14.8/19.8  &\textbf{18.3/23.7}   &3.6/5.4   &33.3   \\
        & GE + class-balanced loss\cite{cui19} &57.0/70.8   &35.1/44.9    &33.2/39.9     &19.1/24.0  &8.4/12.8    &6.1/8.9      &30.0  \\
        & GE + re-weighting    &56.5/70.7   &39.0/49.6     &32.0/38.9     &20.6/25.8      &8.1/12.1    &6.5/9.4  &30.8  \\
        & \textbf{Ours}    &72.1/81.5   &\textbf{50.6/62.6}               &39.9/44.5    &\textbf{26.8/32.8}   &15.2/20.6      &\textbf{10.4/14.4}    &\textbf{38.4}  \\
        \midrule
        \midrule
        \multirow{5}{*}{constrained}
        & GE + focal loss\cite{lin17} &\textbf{64.4/66.8}   &16.7/18.4  &\textbf{35.0/36.0}      &8.7/9.4    &\textbf{18.1/22.9}    &3.5/4.9  &25.4   \\
        &  SMN+TDE\cite{tang20} &46.2/51.4  &25.5/29.1   &27.7/29.9   &13.1/14.9   &16.9/20.3   &8.2/9.8  &24.4 \\
        & GE + class-balanced loss\cite{cui19}  &43.4/48.1   &29.7/33.6   &24.9/26.8   &15.9/17.9   &8.4/12.6    &6.0/8.8   &23.0  \\
        & GE + re-weighting    &40.4/44.6   &32.1/35.9  &22.4/24.2  &16.5/18.3   &8.1/11.9    &6.5/9.3  &22.5  \\
        & \textbf{Ours}    &50.8/52.6   &\textbf{35.2/37.8} &27.6/28.4    &\textbf{18.6/19.6}  &14.6/18.6      &\textbf{9.5/11.7}  &\textbf{27.1}  \\
        \bottomrule
    \end{tabular}
    \label{tab:debiasing}
\end{table*}
\begin{figure}
    \centering
    \includegraphics[scale=0.26]{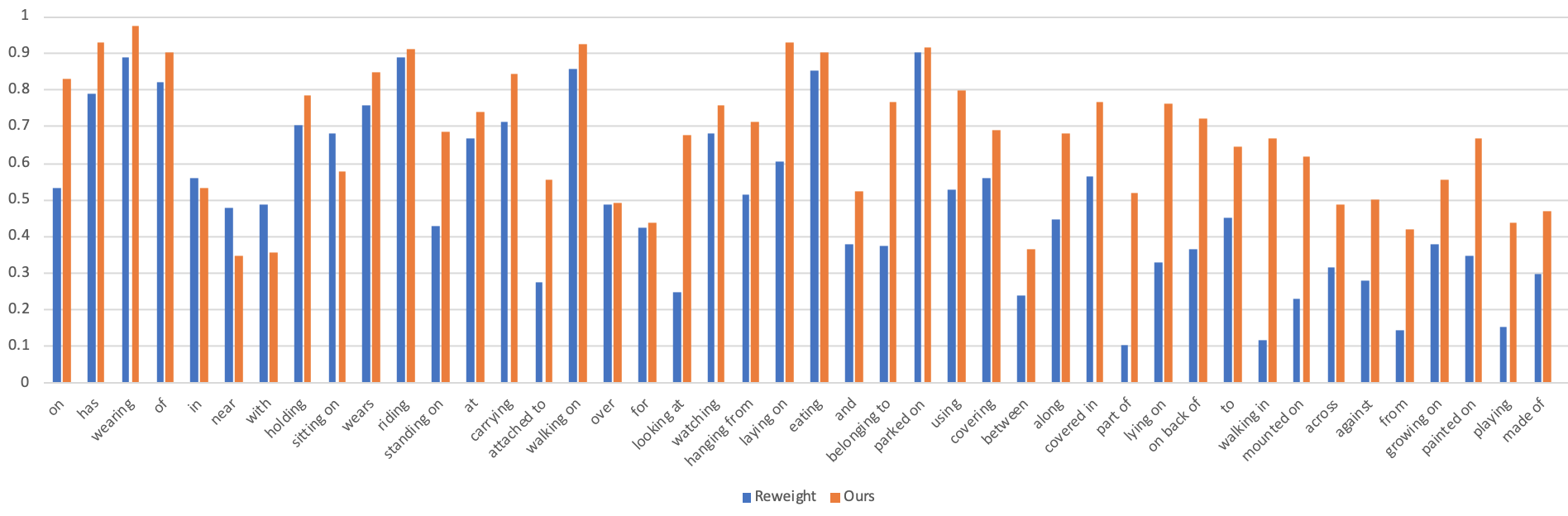}
    \caption{Performance comparison between re-weighting and our method on the VG150 dataset. The unconstrained R@100 for each predicate class on the PredCls task is presented. }
    \label{fig:comparison}
\end{figure}
\begin{table*}[h]
    \centering
    \caption{Performance comparison of different compositions of PCPL on VG150. The constrained and unconstrained mR@50/100 in \% on PredCls, SGCls and SGGen tasks are presented.}
    \begin{tabular}{ccccccccc}
        \toprule
        \multirow{2}{*} & \multirow{2}{*}{Methods} & \multicolumn{2}{c}{PredCls} & \multicolumn{2}{c}{SGCls} & \multicolumn{2}{c}{SGGen} \\
        &  &  mR@50 & mR@100 & mR@50 & mR@100 & mR@50 & mR@100 & Mean\\
        \midrule
        \multirow{4}{*}{unconstrained} & LearntCenter + SoftmaxNorm   &34.1  &45.8  &18.7  &24.5  &4.4    &6.7   &22.4   \\
        & LearntCenter + ScalingNorm &37.2   &49.1    &20.6     &26.4  &5.1    &7.4      &24.3  \\
        & AvgCenter + MinMaxNorm  &49.7   &61.9     &25.4     &31.8      &9.2    &12.0  &31.7  \\
        & \textbf{LearntCenter + MinMaxNorm}    &\textbf{50.6}   &\textbf{62.6}               &\textbf{26.8}    &\textbf{32.8}   &\textbf{10.4}      &\textbf{14.4}    &\textbf{32.9}  \\
        \midrule
        \midrule
        \multirow{4}{*}{constrained} & LearntCenter + SoftmaxNorm  &17.4   &18.9  &9.1      &9.7    &3.9    &5.3  &10.7   \\
        & LearntCenter + ScalingNorm  &19.0   &20.5   &10.1   &10.7   &4.5    &5.8   &11.8  \\
        & AvgCenter + MinMaxNorm    &34.1   &36.9  &17.8  &18.9   &8.6    &10.6  &21.2  \\
        & \textbf{LearntCenter + MinMaxNorm}    &\textbf{35.2}   &\textbf{37.8} &\textbf{18.6}    &\textbf{19.6}  &\textbf{9.5}      &\textbf{11.7}  &\textbf{22.1}  \\
        \bottomrule
    \end{tabular}
    \label{tab:abalation1}
\end{table*}
\begin{table*}[htbp]
    \centering
    \caption{Ablation study of our method.The constrained and unconstrained mR@50/100 in \% on PredCls, SGCls and SGGen tasks are presented.}
    \begin{tabular}{cccccccccc}
        \toprule
        \multirow{2}{*} & \multirow{2}{*}{Methods} & \multicolumn{2}{c}{PredCls} & \multicolumn{2}{c}{SGCls} & \multicolumn{2}{c}{SGGen} \\
        &  &  mR@50 & mR@100 & mR@50 & mR@100 & mR@50 & mR@100 & Mean\\
        \midrule
        \multirow{3}{*}{unconstrained} 
        &  GE  &32.7 &44.0 &18.3 &23.8 &8.3 &11.6 &23.1  \\
        &  GE+PCPL &50.1 &61.9 &26.1 &32.3 &10.1 &14.2 &32.5  \\
        &  \textbf{Ours}   &\textbf{50.6}   &\textbf{62.6}  &\textbf{26.8}    &\textbf{32.8}   &\textbf{10.4} &\textbf{14.4}    &\textbf{32.9}  \\
        \midrule
        \midrule
        \multirow{3}{*}{constrained} 
        &  GE  &17.3 &18.7 &9.3 &9.8 &5.5 &6.5 &11.2  \\
        &  GE+PCPL &34.5 &37.4 &18.1 &19.2 &9.3 &11.2 &21.6  \\
        &  \textbf{Ours}  &\textbf{35.2}   &\textbf{37.8}                &\textbf{18.6}    &\textbf{19.6}  &\textbf{9.5}      &\textbf{11.7}      &\textbf{22.1}  \\
        
        \bottomrule
    \end{tabular}
    \label{tab:abalation2}
\end{table*}
\subsection{Compared methods}
In this section, we first perform a thorough comparison between our proposed method and the existing state-of-the-art methods of scene graph generation, including Iterative Message Passing (IMP+)\cite{xu17}, Frequency baseline (FREQ)\cite{zellers18}, Stacked Motif Networks (SMN)\cite{zellers18}, Knowledge-Embedded Routing Network (KERN)\cite{chen19}, Visual Contexts Tree (VCTree)\cite{tang19} and Stacked Motif Networks with TDE (SMN + TDE) \cite{tang20}.
As shown in Table.~\ref{tab:state-of-the-art}, we present the unconstrained and constrained mR@K on three tasks on the VG150 benchmark. KERN, which explicitly uses the statistical co-occurring prior outperforms SMN by 5.9\% and 2.7\% of the mean mR while VCTree, which mines the implicit relatedness between object pairs by learning a score matrix, gains a slight improvement over KERN.
Though these methods achieve significant progress, TDE, as state-of-the-art on mR@K, is the first method focusing on the long-tailed trouble of SGG. While TDE is a prediction strategy, our proposed PCPL is a training scheme. Our method achieves higher constrained mR@K than others on all three tasks, gaining the mean mR of 22.1\% and 32.9\%, with a relative improvement of 31.5\% and 24.2\% compared with the previous state-of-the-art methods (i.e., SMN+TDE and KERN). Our model evidently outperforms others on PredCls and SGCls in the unconstrained mR@K metric, only slightly lower than KERN on the SGGen task, principally due to the statistical co-occurrence of objects KERN uses to promote the performance of object detection which is not the main concern of our discussion.
\\\indent Secondly, we perform a more in-depth comparison between our method and several commonly used class imbalance handling strategies as well as the previous state-of-the-art debiasing method of SGG (SMN+TDE) to further demonstrate the effectiveness of PCPL. We retrain our baseline model (GE) with focal loss\cite{lin17}, class balanced loss\cite{cui19} and weighted cross entropy loss respectively. The results of R@50/100 and mR@50/100 on three tasks are listed in Table.~\ref{tab:debiasing}. Focal loss, which assigns larger training costs to hard samples, leads to a decline of mR@50/100 from GE. Re-balancing methods, i.e. class-balanced loss and re-weighting gain significant improvements on mR@50/100 but occur huge decrease on R@50/100, indicating over-fitting to tail classes. SMN+TDE\cite{tang20} achieves a relatively balanced performance, promopting the mR@50/100 while preferablely keeping the performance of R50/100. Our method gains further increase on mR@50/100 from re-weighting and acquire comparable R@50/100 with SMN+TDE, though their detector is equipped with resnet\cite{resnet}, which is a more powerful backbone than the VGG16 we use. The comparison suggests that our proposed PCPL can supervise the model to learn a more unbiased representation of scene graph. 

Fig.~\ref{fig:comparison} presents a comparison between our method and re-weighting of the detailed recall@100 of PredCls task on each predicate class ranking by sample frequencies. PCPL performs better than re-weighting on almost all the predicate classes. While evidently promoting the performance of tail classes with few training samples like \emph{walking in}, \emph{mounted on} and \emph{painted on}, PCPL obtains recall@100 on the four head predicates, which accounting of nearly 70\% of the training data, as 83\%, 93\%, 97.5\% and 90.2\%, with improvements of 29.9\%, 13.8\%, 8.6\% and 8.1\% over those of re-weighting. Re-balancing strategies blindly restrain the training of head classes and encourage tail predicates while disregarding the correlations between them, gaining unworthy improvements on tail classes at the cost of massive decrease of the results on head predicates (e.g., \emph{on}). On the contrary, PCPL adaptively assigns optimal loss weights appropriately inversed with their relatedness to predicate classes during the training process. The performance of predicates with weak correlations (e.g., \emph{looking at},\emph{belonging to} and \emph{playing}) improves on account of higher training costs. Although predicate classes having strong correlations with multi classes (e.g., \emph{on}) are assigned with relatively smaller loss weights, their learning benefits from the training process of correlated ones (e.g., \emph{parked on},\emph{standing on} and \emph{walking on}), thus we are able to obtain relatively unbiased results on all predicates. 
\subsection{Ablation Study}
We consider several ablations in Table.~\ref{tab:abalation1} and Table.~\ref{tab:abalation2}. Table.~\ref{tab:abalation1} reveals the different ways to obtain class center representations (i.e., AvgCenter and LearntCenter, using the average of all features of a class to represent the center in every epoch and learning a class center end-to-end,respectively) and normalize the correlation factor(i.e., SoftmaxNorm, ScalingNorm and MinMaxNorm, using softmax function, divided with the maximum value and employing Eq.~\ref{eq:normalize} to obtain correlation factor $\tau$ from global correlation \textbf{u}, respectively). Results show that the composition we use (i.e., LearntCenter + MinMaxNorm) acquires best performance. An explanation for the low performance of SoftmaxNorm and ScalingNorm is that the global correlation \textbf{u} of each predicate is roughly at the same scale, causing the distribution of correlation factor $\tau$ obtained using SoftmaxNorm or ScalingNorm is too smooth to make enough impact on the loss weights while MinMaxNorm magnifies the difference.

The contributes of this paper can be summarized as PCPL, the graph encoder and the noisy label dropping method. To better verify the effectiveness of each components, we perform an ablation study as listed in Table.~\ref{tab:abalation2}. The performance of model with PCPL on all three tasks occurs an evident rise from GE, which clearly shows that our proposed PCPL greatly improves the generalization ability of the model. Meanwhile, GE still markedly outperforms IMP+, FREQ and SMN, indicating the effectiveness of the graph encoders in encoding context and extracting better visual features. Equipped with the noisy label dropping schema, the performance of model gains a sight further improvement, demonstrating its efficiency.

\section{Conclusion}
In this paper, we discover that the key challenge for generating unbiased scene graph lies in the complex relatedness among predicate classes. Thus, we propose a novel PCPL framework which can adaptively assign optimal loss weights to predicates by directly perceiving and explicitly utilizing the correlations among classes. PCPL is further equipped with a graph encoder module to better extract context features. Extensive experiments on the benchmark VG150 dataset show that PCPL performs markedly better on tail classes while well-preserving the performance on head ones, which significantly outperforms previous state-of-the-art methods in mean recall evaluation metric, demonstrating its effectiveness in removing the long-tailed bias of SGG.

\begin{acks}
This paper is partially supported by NSFC (No.31627802) and the
Fundamental Research Funds for the Central Universities.
\end{acks}

\bibliographystyle{ACM-Reference-Format}
\balance
\bibliography{reference}

\end{document}